\documentclass{article}

\usepackage{microtype}
\usepackage{graphicx}
\usepackage{subcaption}
\usepackage{booktabs} 
\usepackage{fancyhdr}
\usepackage{amsfonts,epsfig}
\usepackage{afterpage}
\usepackage{amsmath,amssymb,amsthm, mathtools}
\usepackage{epsf}
\usepackage{float}
\usepackage{multirow}
\usepackage{mathabx}
\usepackage{dsfont}
\usepackage{amsfonts}
\usepackage[textsize=tiny, textwidth=1in, disable]{todonotes}
\usepackage{pgf}
\usepackage{graphicx}
\usepackage{caption}
\usepackage{xcolor}
\usepackage{xspace}
\usepackage{enumitem}
\usepackage{hyperref}
\usepackage{mymacros}

\newcommand\tian[1]{{\color{orange}\textbf{TL: }{#1} }}

\usepackage[accepted]{icml2021}

\icmltitlerunning{Heterogeneity for the Win: One-Shot Federated Clustering}

\begin{document}

\twocolumn[ 
\icmltitle{Heterogeneity for the Win: One-Shot Federated Clustering} 


\icmlsetsymbol{equal}{*}

\begin{icmlauthorlist}
\icmlauthor{Don Kurian Dennis}{to}
\icmlauthor{Tian Li}{to}
\icmlauthor{Virginia Smith}{to}
\end{icmlauthorlist}

\icmlaffiliation{to}{Carnegie Mellon University, Pittsburgh, PA, USA}

\icmlcorrespondingauthor{Don Dennis}{dondennis@cmu.edu}

\icmlkeywords{Machine Learning, ICML}

\vskip 0.3in
]



\printAffiliationsAndNotice{}  

\begin{abstract}
  In this work, we explore the unique challenges---and opportunities---of
unsupervised federated learning (FL).  We develop and analyze a one-shot
federated clustering scheme, \kfed, based on the widely-used Lloyd's method for
$k$-means clustering.  In contrast to many supervised problems, we show that
the issue of {statistical heterogeneity} in federated networks can in fact
benefit our analysis.  We analyse \kfed under a {center separation} assumption
and compare it to the best known requirements of its centralized counterpart.
Our analysis shows that in heterogeneous regimes  where the number of clusters
per device $(k')$ is smaller than the total number of clusters over the network
$k$, $(k'\le \sqrt{k})$, we can use heterogeneity to our
advantage---significantly weakening the cluster separation requirements for
\kfed.  From a practical viewpoint, \kfed also has many desirable properties:
it requires only one round of communication, can run asynchronously, and can
handle partial participation or node/network failures. We motivate our analysis
with experiments on common FL benchmarks, and highlight the practical utility
of one-shot clustering through use-cases in personalized FL and device
sampling.

\end{abstract}
\section{Introduction}

Federated learning (FL) aims to perform  machine learning over large,
heterogeneous networks of devices such as mobile phones or
wearables~\cite{mcmahan2017communication}.  While significant attention has
been given to the problem of supervised learning in such settings, the problem
of unsupervised federated learning has been relatively
unexplored~\cite{kairouz2019advances}.  In this work, we show that unsupervised
learning presents unique opportunities for FL, specifically for the task of
clustering data that resides in a federated network. 

Clustering is a crucial first step in many learning tasks.  In the case of
federated learning, clustering has found applications in
client-selection~\cite{cho2020client}, personalization~\cite{ghosh2020} and
exploratory data analysis.  While many works have explored techniques for
distributed clustering (Section~\ref{sec:background}), most do not take into
account the unique challenges of federated learning, such as {statistical
heterogeneity}, systems heterogeneity,  and stringent {communication
constraints}~\cite{li2020federated}\footnote{Privacy, while an important
concern for many federated applications, is not the main focus of our work.
However, a possible benefit of the one-shot nature of \kfed is that it
requires significantly fewer messages to be shared over the network relative to
standard iterative techniques such as distributed $k$-means.}.  These
challenges can complicate analyses, reduce  efficiency, and lead to practical
issues with stragglers and device failures.  In this work, we study
communication-efficient distributed clustering in settings where the data is
non-identically distributed across the network (i.e., heterogeneous), and
devices can join and leave the network abruptly.  For such settings, we develop
and analyse a one-shot clustering scheme, \kfed, based on the classical
Lloyd's heuristic~\cite{lloyd01} for clustering.

The method we propose, \kfed, requires only one round of communication with a
central server. Each device, indexed by $z$, solves a local $k^{(z)}$-means
problem and then communicates its local cluster means via a message of size
$O(dk^{(z)})$.  As we show in Section 3, this  allows for device failures, only
requiring that there are enough devices available in the network such that $k$
target clusters exist in the data. Moreover, it is possible to cluster points
in previously unavailable devices via a simple recomputation at the central
server.

Beyond the practical benefits of \kfed, our work is unique in rigorously
demonstrating a problem setting where possible benefits of statistical
heterogeneity exist for federated learning.  In particular, in supervised
learning, many works have highlighted detrimental effects of statistical
heterogeneity, observing that heterogeneity can lead to poor convergence for
federated optimization methods~\cite{mcmahan2017communication,li2018federated},
result in unfair models~\cite{mohri2019agnostic}, or necessitate novel forms of
personalization~\cite{smith2017federated,mansour2020three}. In contrast to
these works, we show that for the specific notion of heterogeneity considered
herein (provided in Definition~\ref{def:heterogeneity} and motivated by the
application of clustering), heterogeneity can in fact have measurable benefits
for our approach.

More specifically, similar to many works in clustering~(\citet{kumarkannan01,
awasthisheffet01} and references therein), we analyse \kfed under a
center-separation assumption; that is, we assume that the mean of the
clusters are well separated. We also consider a specific notion of
heterogeneity: given a target clustering with $k$ clusters that we wish to
recover from the data, we assume that each device contains data from only $k'
\le \sqrt{k}$ of these target clusters. For instance, for clustering data
generated by a mixture of $k$ well separated Gaussians, we assume that each
device contains data from $k' \le \sqrt{k}$ component Gaussians.  In this
regime, we show that our separation  requirement is similar to that of the
centralized counterpart. Further, while the centralized setting
requires all pairs of cluster centers to satisfy a $\Omega(\sqrt{k})$ center
separation requirement, the federated approach can handle a large
fraction of cluster pairs only satisfying a weaker $\Omega(k^\frac{1}{4})$
separation requirement.  This is the first result we are aware of that analyzes
the benefits of heterogeneity in the context of federated
clustering.

\vspace{.5em}
\textbf{Contributions.} We propose and analyze a one-shot communication scheme
for federated clustering.  Our proposed method, \kfed, addresses common
practical concerns in federated settings, such as high communication costs,
stragglers, and device failures.  Theoretically, we show that \kfed~performs
similarly to centralized clustering in regimes where each device only has data
from at most $\sqrt{k}$ clusters with a similar $\Omega(\sqrt{k})$ center
separation requirement.  Moreover, in contrast to the centralized setting, we
show that a large number of cluster pairs need only a $\Omega(k^\frac{1}{4})$
weaker separation assumption in heterogeneous networks, thus allowing a broader
class of problems to be solved in this setting compared with centralized
clustering.  We demonstrate our method through experiments on common FL
benchmarks, and explore the applicability of \kfed to problems in personalized
federated learning and device sampling.  Our work highlights that
heterogeneity can have distinct benefits for a subset
of problems in federated learning.
\vspace{.5em}

\section{Background and Related Work}
\label{sec:background}

\textbf{Centralized Clustering.} Clustering is one of the most widely-used
unsupervised learning tasks, and has been extensively studied in both
centralized and distributed settings. Although a variety of clustering methods
exist, Lloyd's heuristic~\cite{lloyd01} remains popular due in part to its
simplicity. In Lloyd's  method, we start with an initial set of $k$ centers. We
then assign each point to its nearest center and reassign the centers to be
the mean of all the points assigned to it, continuing this process till
termination.  While it is easy to show that this method terminates, it is also
known that this process can take superpolynomial time to
converge~\cite{arthur01slow}.  However, under suitable assumptions and careful
choice of the initial centers, it can be shown to converge in polynomial
time~\cite{arthur01slow, ostrovsky01, kumarkannan01, awasthisheffet01}. 

The method we propose, \kfed (Section~\ref{sec:fedtheory}), is a simple,
communication-efficient distributed variant of these classical techniques.
\kfed runs a variant of Lloyd's method for $k$-means clustering locally on each
device, and then performs one round of communication to aggregate and assign
clusters.  Our work builds on the analysis of a variant of Lloyd's algorithm
developed by~\citet{kumarkannan01} and later improved in
\citet{awasthisheffet01} for the problem of clustering data from mixture
distributions and other related results
~\citep[e.g.,][]{mcsherry2001spectral,ostrovsky01}. These works develop a
deterministic framework with no generative assumptions on the data. Our
analysis follows this framework and does not make any generative assumptions on
the data. 

\vspace{.5em}
\textbf{Parallel and Distributed Clustering.}  Many works have explored
parallel or distributed implementations of centralized clustering
techniques~\cite{dhillon2002data,tasoulis2004unsupervised,dutta2005k,
bahmani2012scalable,xu1999fast}.  Unlike the one-shot communication scheme
explored herein, these methods are typically direct parallel implementations of
methods such as Lloyd's heuristic or DBSCAN~\cite{ester1996density}, and
require numerous rounds of communication.  Another line of work has considered
communication-efficient distributed clustering variants that require only one
or two rounds of communication~\citep[e.g.,][]{kargupta2001distributed,januzaj2004dbdc,
feldman2012effective,balcan2013distributed,bateni2014distributed,
bachem2018scalable}.  These works are mostly empirical, in that there are no
provable guarantees on the approximation quality of the distributed schemes;
the works
of~\citet{balcan2013distributed,bateni2014distributed,bachem2018scalable}
differ by providing communication-efficient distributed coreset methods for
clustering, along with provable approximation guarantees. However, these works
do not explore the federated setting or potential benefits of heterogeneity.

\vspace{.5em}
\textbf{Federated Clustering.} 
Several works have explored clustering in the context of supervised FL as a way
to better model non-IID
data~\cite{smith2017federated,ghosh2019robust,ghosh2020,sattler2020clustered}.
These works differ from our own by  clustering specifically in terms of
devices, focusing on the downstream supervised learning task, and using either
iterative~\cite{smith2017federated,ghosh2020,sattler2020clustered} or
centralized~\cite{ghosh2019robust} clustering schemes.  Though not the main of
focus of our work, in Section~\ref{sec:exp} we demonstrate the applicability of
one-shot clustering by showing how \kfed can be used as a simple pre-processing
step to deliver personalized federated learning---achieving similar or superior
performance relative to the recent iterative approach for clustered FL proposed
in~\citet{ghosh2020}. 

More recently, a distributed matrix factorization based clustering approach was
explored in~\citet{wang2020} for the purposes of unsupervised learning.
However, while the authors consider the impact of statistical heterogeneity on
their convergence guarantees, the focus is not on one-shot clustering or on
showing distinct benefits of heterogeneity in their analyses.

\section{\kfed: Preliminaries and Main Results}
\label{sec:method}

In this section, we begin by discussing some preliminaries and existing results
in clustering related to Lloyd-type methods. In Section~\ref{sec:centralk}, we
present the deterministic framework of~\citet{awasthisheffet01} for centralized
clustering, which we build upon.  We present our method \kfed and state our
theoretical results in Section~\ref{sec:fedtheory}. We provide detailed proofs
in Appendix~\ref{app:proofs}.

\subsection{Centralized $k$-means}
\label{sec:centralk}

In the standard (centralized) $k$-means problem, we are given a matrix ${A} \in
\R^{n\times d}$ where each row $A_i$ is a data point in $\R^d$. We are also
given a fixed positive integer $k \le n$, and our objective is to partition the
data points into $k$ disjoint partitions, $\mathcal{T} = (T_1, \dots, T_k),$ so
as to minimize the $k$-means cost:
\begin{equation}
  \phi(\mathcal{T}) = \sum_{j=1}^k \sum_{i \in T_j}\ltwo{A_i-\mu(T_j)}^2 \,.
\end{equation}
Here we use $\mu(S)$ as an operator to indicate the mean of the points indexed
by $S$, i.e., $\mu(S) = \frac{1}{\abs*{S}}\sum_{i \in S} A_i$.  To ease
notation, we simplify this as  $\mu_r := \mu(T_r)$, when $T_r$ is
unambiguous.

\begin{algorithm}[tb]
  \caption{Local $k^{(z)}$-means \cite{awasthisheffet01}}
\label{algodevice}
\begin{algorithmic}[1]
  \STATE {\bfseries Input:} On device indexed by $z$, the  matrix of data
  points $A^{(z)}$, integer $k^{(z)}$;
  \STATE Project $A^{(z)}$ onto the subspace spanned by the top $k^{(z)}$
  singular vectors to get $\hat A^{(z)}$. Run any standard $10$-approximation
  algorithm on the projected data and estimate $k^{(z)}$ centers ($\nu_1,
  \nu_2, \dots, \nu_{k^{(z)}}$).
  \STATE Set 
  \[
    S_r \leftarrow \set{i:\ \ltwos{\hat A^{(z)}_i -\nu_r} \le \frac{1}{3}
    \ltwos{\hat A^{(z)}_i - \nu_s},\text{ for every } s}
  \] and $\theta^{(z)}_r \leftarrow \mu(S_r)$
  \STATE Run Lloyd steps until convergence
  \[
    U^{(z)}_r\leftarrow \set{i:\ \ltwos{A_i^{(z)} - \theta^{(z)}_r} 
      \le \ltwos{A_i^{(z)} -\theta^{(z)}_s}, \forall s}
    \] and $\theta^{(z)}_r \leftarrow \mu(U^{(z)}_r)$.
  \STATE {\bfseries Return:} Cluster assignments $(U^{(z)}_1, U^{(z)}_2, \dots,
  U^{(z)}_{k^{(z)}})$ and their means $\Theta^{(z)} = (\theta^{(z)}_1, \dots,
  \theta_{k^{(z)}})$.
\end{algorithmic}
\end{algorithm}

While the \kmeans problem as stated here does not specify any generative model
for the data points $A_i$, a popular setting to consider is when the data is
sampled from a mixture of $k$-distributions in $d$-dimensions ($k \ll d$). For
instance, we could imagine the data points as being sampled from a mixture of
$k$ Gaussian distributions. This generative model also introduces a notion of a
\textit{target clustering}, $\mathcal{T} = (T_1, \dots, T_k)$ where the set
$T_i$ contains all points generated by the $i$-th component distribution. Many
distribution dependent results are known for the problem of clustering
distributions (see~\citet{kumarkannan01}). In general, they can be stated as:
If the means of the distributions are $\text{poly}(k)$ standard deviations
apart, then we can cluster the data in polynomial time.  \citet{kumarkannan01}
introduce a deterministic (distribution independent) framework that
encompasses many of these known results. This work was later simplified and
improved by \citet{awasthisheffet01}. We state the main results of this framework
here, after stating the notation we use. We emphasis that in our analysis we
make no assumptions on how the data is generated; all relevant quantities  only
depend on the provided data.

\textbf{Notation.} We now introduce several definitions and notations that will
be used throughout the paper. Let $\ops{A}$ denote the spectral norm of a
matrix $A$, defined as $\ops{A} = \max_{u:\ltwos{u}=1} \ltwos{Au}$, and let
$\norm*{A_i}_2$ denote the $\ell_2$ norm of a vector $A_i$. For consistency, we
index individual rows of $A$ with $i$ and $j$. Moreover, when a target
clustering $T_1, \dots, T_k$ is fixed, we index clusters with $r, s$, e.g.,
$A_r$ is the matrix of points indexed by $T_r$.  For notational convenience, we
let $c(A_i)$ to denote the cluster index for data point $A_i$ such that, $A_i
\in T_{c(A_i)}$. For some set of points $M$, and another point say $x$, let
$d_M(x)$ denote the distance of $x$ to the set $M$, defined as $d_M(x) =
\min_{y \in M} \ltwos{x - y}$. Finally, let $C$ be a $n \times d$ matrix with
each row $C_i = \mu_{c(A_i)}$. For  cluster $T_r$ with $n_r = \abs*{T_r}$, we
define
\begin{equation}
  \tilde\Delta_{r} := \sqrt{k}\frac{\ops{A-C}}{\sqrt{n_r}}.
\end{equation}
Here the quantity $\ops{A-C}/\sqrt{n_r}$ can be thought of as a deterministic
{analogue} of the standard deviation; it measures the maximum average variance
along any direction. Thus instead of reasoning about the separation between two
clusters $T_r$ and $T_s$ in terms of the standard deviation, we will use
$(\tilde\Delta_r + \tilde\Delta_s)$.  In particular, we say that the two
clusters $T_r$ and $T_s$ are \textit{well separated} if for large enough
constant $c$, their means satisfy:
\begin{align}
  \ltwo{\mu_r - \mu_s} \ge c(\tilde\Delta_r + \tilde\Delta_s) \, .
  \label{eq:separation}
\end{align}  
Again, we can interpret this as saying that two clusters are well separated if
their means are $c$-standard-deviations apart.\footnote{ Any $c \ge 100$ is
sufficient for our arguments (see Lemma~\ref{lemma:localbound}).} Using the center separation assumption
in~\eqref{eq:separation}, \citet{awasthisheffet01} show that for a target
clustering $T_1, T_2, \dots, T_k$ satisfying the separation assumption, the
variant of Lloyd's algorithm presented in~\algoA when applied to the
centralized clustering problem correctly clusters all but a small fraction of
the data points. We state their result formally in Lemma~\ref{lemma:awasthi},
but before that we define a \textit{proximity condition}, that will be used to
precisely characterize the misclassified points.
\begin{definition}\label{def:proxcondition} A point $A_i$ for some $i \in T_s$
  is said to satisfy the {proximity condition}, if for every $r \ne s$, the
  projection of $A_i$ onto the line connecting $\mu_r$ and $\mu_s$, denoted by
  $\bar A_i$ satisfies
  \[
    \ltwo{\bar A_i - \mu_r} - \ltwo{\bar A_i - \mu_s} 
    \ge \left(\frac{1}{\sqrt{n_r}} + \frac{1}{\sqrt{n_s}}\right)\ops{A-C}.
  \]
\end{definition} Thus a point $A_i$ for $i\in T_s$ satisfies the proximity
condition if its projection on the line connecting $\mu_r$ and $\mu_s$ is
closer to $\mu_s$ by $\ops{A-C}(\frac{1}{\sqrt{n_r}} + \frac{1}{\sqrt{n_s}})$.
We refer to points that do not satisfy the proximity condition as `bad points'.
We now state the main result from~\citet{awasthisheffet01} in the following
lemma.

\begin{lemma}[Awasthi-Sheffet, 2011]\label{lemma:awasthi} Let $\mathcal{T} =
  (T_1, \dots, T_k)$ be the target clustering. Assume that each pair of
  clusters $T_r$ and $T_s$ are \textit{well separated}. Then,  after step 2
  of~\algoA, for every $r$, it holds that ${\ltwos{\mu(S_r)-\mu_r} \le
  \frac{25}{c}\frac{1}{\sqrt{n_r}}\ops{A-C}}$. Moreover, if the number of bad
  points is $\eps n$, then (a) the clustering $\set{U_1, U_2, \dots, U_k}$
  misclassifies  no more than $(\eps + O(1)c^{-4})n$ points and (b) $\eps <
  O((c-\frac{1}{\sqrt{k}})^{-2})$. Finally, if $\eps = 0$ then all points are
  correctly assigned.
\end{lemma} When we say misclassify, we mean with respect to $\mathcal{T}$ and
up to a permutation of labels.  Lemma~\ref{lemma:awasthi} tells us that the
cluster means, $\mu(S_r)$, are not very far away from the target cluster means,
$\mu_r$. Note that there are no distribution dependent terms in this statement;
all relevant quantities are defined in terms of the data matrix $A$ and
$\mathcal{T}$.

\subsection{\kfed: Method and Main Result }
\label{sec:fedtheory}

We now turn our attention to clustering data in a federated network.
In our setting, we assume that all the devices in the network can communicate
with a central server.  Our clustering method \kfed, described
in Algorithm~\ref{algo:kfed},   can be thought of as working in two stages.
In the first stage, each device solves a local clustering subproblem and
computes the cluster means for this subproblem. In the second stage, the
central server accumulates and aggregates the results to compute the final
clustering.

\textbf{Notation.} Let $A$ be an $n \times d$ data matrix of all the data
points in our network. We index individual devices by $z \in [Z]$ and thus, we
denote the data-matrix for any particular device by ${A}^{\devz} \in
\R^{n^\devz \times d}$, where $n^{\devz}$ is the number of data points on the
device. Let $n_{\min} = \min_{z} n^\devz$.
Note that $A^{\devz}$ is some subset of rows of $A$. Let $\mathcal{T} = (T_1,
\dots, T_k)$ be a clustering of all the data, referred to as a target
clustering. For a fixed $\mathcal{T}$, let $\mathcal{T}^{(z)} = (T_1^{(z)},
T_2^{(z)}, \dots, T_k^{(z)})$ be subsets of our target clustering that reside
on a device $z$.  Note that some $T^{(z)}_r$ could be empty. Let $k^{(z)}$ be
the number of non-empty subsets on device $z$ and let  $k' = \max_z k^{(z)}$.
Our notion of heterogeneity is formally defined based on the value of $k'$, as
described below.

\vspace{.5em}
\begin{definition}[\textbf{Heterogeneity of
  Clustering}]\label{def:heterogeneity} In the context of clustering, we say
  that a federated network with sufficient data is heterogeneous if $k' \leq
  \sqrt{k}$. The lower the ratio between $k'$ and $\sqrt{k}$, the more
  heterogeneity exists in the network.
\end{definition}

Intuitively, this definition of heterogeneity states that---in contrast to the
data from the $k$ total clusters being partitioned in an IID fashion across the
network---the data are partitioned in an non-IID fashion, such that only data
from  a small number of clusters (at most $k'$) exists on each device. Such
non-IID partitioning is reasonable to expect in heterogeneous federated
networks with a large number of clusters, since the distribution of data on
each device may differ, and it is not possible to actively re-distribute data
across the network. For instance, consider identifying interests of mobile
phone users based on the interaction data on an application.  Here the
interaction data is generated by the user on their particular device, and will
reflect the tastes of individual. While the total number of ‘tastes’ (clusters)
over the entire network could be quite large, a typical user will be interested
in only a small number of them. With this definition in mind, we next describe
our one-shot clustering method, \kfed, and analyze it in heterogeneous
regimes.

\paragraph{Method Description.} Similar to the
centralized case (Section~\ref{sec:centralk}), let $C^{(z)}$ be a $n^{(z)}
\times d$ matrix of the local cluster means, i.e. of $\mathcal{T}^{(z)}$.
Consider a non-empty susbset $T^{(z)}_r$ of cluster $T_r$ on some device and
let $n^{(z)}_r = \abs*{T^{(z)}_r}$. We assume that there is a constant $m_0 >
1$, such that $n^{(z)}_r \ge \frac{1}{m_0}n_r$ for all $r$.  We will use this
quantity to ensure that individual devices have `enough' points.  Let,
\begin{align}
  \Delta_r = k'\frac{\ops{A-C}}{\sqrt{n_r}},
  \quad\text{and}\quad
  \lambda = \sqrt{k'}\left(\frac{\ops{A-C}}{\sqrt{n_{\min}}}\right)\, .
  \label{def:separations}
\end{align} 

In the first step of \kfed (\algoB), each (available) device $z \in [Z]$
runs~\algoA locally and solves a local clustering problem with their local
dataset $A^{(z)}$ and parameter $k^{(z)}$. We assume that $k^{(z)}$ is known.
This stage outputs \textit{device cluster centers} $\Theta^{(z)} =
(\theta_1^{(z)}, \dots, \theta^{(z)}_{k^{(z)}})$ and cluster assignments,
$U^{(z)}_1, \dots, U^{(z)}_{k^{(z)}}$ for each device $z$.  At this stage, note
that even though each device has classified its own points into clusters, we do
not yet have a clustering for points across devices.  The central server
attempts to create this clustering by aggregating the device cluster centers
and separating them into $k$ sets, $\tau_1, \dots, \tau_k$.  These sets
\textit{induce a clustering} of the data on the network as defined here:

\begin{definition}[\kfed induced clustering]\label{def:induced-clustering} Let
  $\tau_1, \tau_2, \dots, \tau_k$  be the clustering of device centers returned
  by Algorithm~\ref{algo:kfed}. Define,
  \[
    T'_r = \set{i: A^{(z)}_i \in U^{(z)}_s \text{ and } \theta^{(z)}_s \in
    \tau_r, z \in [Z], s \in [k^{(z)}]}.
  \] Then, $\mathcal{T}' = (T'_1, \dots, T'_k)$ form a disjoint partition of
  the entire data, called the \kfed induced clustering.
\end{definition}

\begin{algorithm}[tb]
 \caption{\kfed}
 \label{algo:kfed}
\begin{algorithmic}[1]
  \STATE On each device $z\in [Z]$, run \algoA with local data $A^{(z)}$ and
  $k^{(z)}$ and obtain device cluster centers $\Theta^{(z)} = (\theta_1^{(z)},
  \dots, \theta_{k^{(z)}}^{(z)})$ at the central node.
  \STATE Pick any $z \in [Z]$ and let $M \leftarrow \Theta^{(z)}$.
  \REPEAT
   \STATE Let $\bar{\theta} \leftarrow \argmax_{z \in [Z], i \in [k]}
   d_M(\theta_i^z)$. That is, the farthest $\theta_i^{(z)}$ from the set $M$.
   \STATE $M \leftarrow M \cup \set{\bar{\theta}}$.
  \UNTIL{there are $k$ points in $M$, i.e. $\abs{M} = k$}
  \STATE Run one round of Lloyd's heuristic to cluster points $\theta^{(z)}_i$,
  $z \in [Z], i \in [k]$ into $k$ sets/clusters, $(\tau_1, \tau_2, \dots,
  \tau_k)$. Use points in $M$ as initial centers.
  \STATE{\bfseries Return:} the clustering $(\tau_1, \tau_2, \dots, \tau_k)$ of
  the device cluster centers and the corresponding \kfed induced clustering
  (Definition \ref{def:induced-clustering}).
\end{algorithmic}
\end{algorithm}

For our analysis comparing the quality of the \kfed induced clustering,
$\mathcal{T}'$, to our target clustering $\mathcal{T}$, we require two
different separation assumptions. We refer to them as \textit{active} and
\textit{inactive} separation and introduce them through the following two
definitions.

\begin{definition}[Active/Inactive cluster pairs] A pair of clusters $(T_r,
  T_s)$ are said to be an active pair if there exists at least one device that
  contains data points from both $T_r$ and $T_s$. If no device has data points
  from both clusters $T_r$ and $T_s$, we refer to the cluster pair $(T_r, T_s)$
  as an inactive pair.
\end{definition}

\begin{definition} We say that two clusters $T_r$ and $T_s$ satisfies the
  active separation requirement if, $\ltwo{\mu_r - \mu_s} \ge
  2c\sqrt{m_0}(\Delta_{r} + \Delta_s)$, for some large enough constant $c$.
  Similarly, we say that they satisfy the inactive separation requirement if
  $\ltwo{\mu_r - \mu_s} \ge 10\sqrt{m_0}(\lambda_r + \lambda_s)$.
\end{definition} 

Intuitively, these notions capture the difficulty in clustering two different
types of clusters pairs---active and inactive cluster pairs.  If no device
has data from both $T_r$ and $T_s$ (i.e. an inactive pair), then the clustering
sub-problems individual devices have to solve is easier since they never
involve data from both of these clusters simultaneously. Thus the separation
requirement for inactive cluster pairs is weaker than that for an active
cluster pair. We now state our main theorem, which characterizes the
performance of \kfed. We provide a detailed proof in Appendix~\ref{app:proofs}.

\begin{theorem}[Main theorem]\label{th:gen} Let $\mathcal{T} = (T_1, T_2,
  \dots, T_k)$  be a fixed target clustering of the data on a federated
  network. Let $m_0>1$ be such that, $\abs*{T^{(z)}_r} \ge
  \frac{1}{m_0}\abs*{T_r}$ for all $r, s$ and for all $z \in [Z]$. Assume that
  each active cluster pairs $T_r$ and $T_s$ satisfy the active separation
  requirement, i.e.,
  \[
    \ltwos{\mu(T_r) - \mu(T_s)} \ge c\sqrt{m_0}(\Delta_r+ \Delta_s).
  \] Further, assume that for each inactive cluster pairs $T_r, T_s$, 
  \[
    \ltwos{\mu(T_r) - \mu(T_s)} \ge 10\sqrt{m_0}\lambda \, .
  \] Then, at termination of~\kfed all but $O(\frac{1}{c^2})n$ points are
  correctly classified. Moreover, if for each device $z$, the data points
  $A^{(z)}$ satisfy the proximity condition
  (Definition~\ref{def:proxcondition}) for its local problem,  then all points
  are classified correctly.
\end{theorem}

As before, by classified we mean that the clustering  $\mathcal{T}'$ produced
by \kfed and $\mathcal{T}$ agree on all but $O(\frac{1}{c^2})n$ points, up to
permutation of labels of $\mathcal{T}$. Note that when $k' \approx k$, our
active separation requirement is stricter than that required in centralized
clustering $(\Omega({k})$ vs $\Omega(\sqrt{k}))$.  Further, as one would
expect, as the number of points per cluster on each device decreases, the local
clustering becomes harder. This is highlighted by our adverse dependency on
$\sqrt{m_0}$.

However, in contrast to the general distributed learning framework where each
device typically has a random subset of the data, the data residing on the
devices in federated networks are typically generated locally and thus the
partition of data among the devices is non-identically distributed.
Specifically, in practice, the number of subsets of target clusters that reside
on a device may be much smaller than the total number of clusters.  Thus, as
outlined in Definition~\ref{def:heterogeneity}, we look at the cases where $k'
\le \sqrt{k}$. Observe that in such settings, our active separation requirement
reduces to that of the centralized $k$-means problem (with an additional
$\sqrt{m_0}$ penalty) and our inactive separation requirement weakens to
$k^{{1}/{4}}$. We state this formally in Corollary~\ref{cor:main}.
\begin{corollary} \label{cor:main} Assuming $k' \le \sqrt{k}$, an active
  cluster pair $(T_r, T_s)$ satisfies the active separation requirement if
  \begin{align*}
    \ltwos{\mu_r - \mu_s} &\ge c\sqrt{m_0
    k}\left(\frac{\ops{A-C}}{\sqrt{n_r}}+\frac{\ops{A-C}}{\sqrt{n_s}}\right)\\
    &= c\sqrt{m_0}(\Delta_r + \Delta_s).
  \end{align*} Similarly, an inactive cluster pair $(T_r, T_s)$ satisfies the
  inactive separation requirement if
  \[
    \ltwos{\mu_r - \mu_s} 
    \ge 10\sqrt{m_0}k^{\frac{1}{4}}
      \left(\frac{\ops{A-C}}{\sqrt{n_r}} + \frac{\ops{A-C}}{\sqrt{n_s}}\right).
  \]
\end{corollary}
Thus in this setting of $k' < \sqrt{k}$, \kfed recovers the target partitions
in only one round of communication. Moreover, inactive cluster pairs need only
satisfy our $\Omega(k^{\frac{1}{4}})$ separation requirement as opposed to the
$\Omega(\sqrt{k})$ separation that all cluster pairs need to satisfy in the
centralized setting for Lemma~\ref{lemma:awasthi} to hold. This highlights that
there exists a benefit of heterogeneity in the context of running \kfed over
federated networks.

\paragraph{Practical benefits of \kfed.} Finally, we highlight several
practical benefits of the \kfed method:
\begin{itemize}
  \item \textit{One-shot:} \kfed only requires one round of communication for
    each device: one outgoing message to send the local clustering results and one incoming message to receive
    cluster identity information.
 \item \textit{No network-wide synchronization:} Classical parallel implementations of
   Lloyd's heuristic and variants~\citep[e.g.,][]{dhillon2002data}, require a
    network wide synchronization/initialization step. Unlike these methods,
    each device in $\kfed$ works independently does not require an
    initialization/synchronization step.
 \item \textit{New devices/Device Failures:} Assuming we have already performed
   clustering on the current network, for any new device entering the network,
    either from a previous failure or as a new participant, computing the
    clustering information can be done without involving any other device in
    the network. As we show in Theorem~\ref{th:runtime} (below), simply
    assigning any new local cluster center $\theta_i^{(z)}$ from the new device
    $z$, to the nearest device cluster mean in $M$ sufficient. The central
    server only has to maintain $k$ cluster means $\mu(\tau_1), \dots,
    \mu(\tau_k)$ to perform this update. 
\end{itemize}

  \begin{theorem}\label{th:runtime} Steps 2-8 of \kfed take $O(Zk'\cdot
    k^2)$ pairwise distance computations to terminate. Further, after the
    set $M$ in Step 6 has been computed, new local cluster centers
    $\Theta^\devz$ from a yet unseen device $z$ can be correctly assigned
    in $O(k'\cdot k)$ distance computations.
  \end{theorem}

As we show in Section~\ref{sec:exp}, these properties of \kfed make it an
ideal candidate for being used as an \textit{inexpensive heuristic} for
clustering in federated networks, either for data exploration or as part of a
preprocessing step for another algorithm, even in settings where the separation requirements
 are not formally satisfied.

\section{Applications and Experiments}
\label{sec:exp}

We now present experimental evaluation of \kfed. We first specialize the theory
to the special case where data is drawn from a mixture of $k$ Gaussians in
Section 4.1 to validate our theory on synthetic data. In Section 4.2, we
evaluate \kfed on real datasets---presenting experimental evidence that highlights
the benefit of heterogeneity and the communication efficiency of \kfed. We further
present two applications of \kfed, in client selection as well as
personalization. The dataset details for each experiment can be found in the
corresponding section. Implementation of \kfed and experimental setup details can be
found at: \url{http://github.com/metastableB/kfed/}.

\subsection{Separating Mixture of Gaussians}

We first specialize our theorem to the case of separating data generated from a
mixture of $k$ Gaussians $F_1, F_2, \dots, F_k$. Let $\mu_r = \mu(F_r)$ be the
mean of the mixture component $F_r$ and let $w_1, w_2, \dots, w_k$ be the
mixing weights.  Finally, let $w_{\min} = \min_{r} w_r$ be the minimum mixing
weight. Let $\sigma_{\max}$ be the maximum variance along any direction among
all the component distributions. Assume this data resides over our devices such
that no single device has data from more than $k' < \sqrt{k}$ components. We
state the following theorem (proved in Appendix~\ref{app:proofs}) that
specifies the conditions required for this setup to satisfy our separation
assumptions:

\begin{theorem}\label{th:mixgauss} Let the total number of data pints, $n =
  {\text{poly}}\left(\frac{d}{w_{\min}}\right)$. Then any active cluster pairs
  $r, s$ satisfy the active separation requirement with high probability if;
  \[
    \ltwos{\mu_r - \mu_s} \ge
    \frac{c\sqrt{k m_0}\sigma_{\max}}{\sqrt{w_{\min}}}
    \text{polylog}\left(\frac{d}{w_{\min}}\right).
   \] Further, an inactive cluster pairs $r', s'$ satisfy the inactive
   separation requirement with high probability if
  \[
    \ltwos{\mu_{r'} - \mu_{s'}} \ge
    \frac{c\sqrt{m_0}k^{\frac{1}{4}}\sigma_{\max}}{\sqrt{w_{\min}}}
    \text{polylog}\left(\frac{d}{w_{\min}}\right).
   \] Finally, with this separation in place, all points satisfy the proximity
   condition with high probability.
\end{theorem}
Concretely, in this setup \kfed recovers the target clustering exactly with
high probability. To empirically evaluate our theory, we instantiate an
simplified  instance of the above setup as follows:

\begin{table}
  \caption{Clustering accuracy for clustering a mixture of Gaussians. Here for all
  instances we choose $k' = \sqrt{k}$. We can see that the one-shot clustering
  produced by \kfed agrees with the target clustering with high accuracy,
  particularly when $k$ is relatively small compared to $d$. }
  \vspace{1em}
  \centering
  \label{table:synthetic}
  \begin{tabular}{c c} 
     \toprule[\heavyrulewidth]
         \textbf{Parameters} & \textbf{Accuracy} \\
        \midrule
        $(d=100, k=16, m_0 = 5, c=100)$ & $100.00 \pm 0.00$  \\
        $(d=100, k=64, m_0 = 5, c=100)$ & $98.82 \pm 0.70$  \\
        $(d=300, k=64, m_0 = 5, c=100)$ & $99.27 \pm 0.73$  \\
        $(d=300, k=100, m_0 = 5, c=100)$ & $98.40 \pm 0.80$  \\
        $(d=300, k=16, m_0 = 5, c=100)$ & $100.00 \pm 0.00$  \\
    \bottomrule[\heavyrulewidth]
  \end{tabular}
\end{table}

\textbf{Setup.}  Again consider the Gaussian components $F_1, \dots, F_k$, and
define the set of integers $G_i = \set{p \mid (i - 1) \times \sqrt{k} \le p \le
i\times \sqrt{k}}$. These sets $G_i$ thus can be used to index the Gaussian
components $(F_{(i-1)\sqrt{k}}, \dots, F_{i\sqrt{k}})$. For each $G_i$,
construct a set of data points $D_i$ by sampling $\text{poly}(dk)$ samples from
each component $F_p$ for $p \in G_i$. Thus the set $D_i$ contains
$\sqrt{k}\cdot\text{poly}(dk)$ samples $(w_r = \frac{1}{k}, \forall r)$. Pick
$m_0$ and for each set of data points $D_i$, distribute the data among $m_0$
devices such that each device receives exactly
$\frac{1}{m_0}\cdot\text{poly}(dk)$ samples. We now run \kfed on this setup and
measure the quality of the clustering averaged over 10 runs, (shown in
Table~\ref{table:synthetic}). As one would expect, the clustering produced by
\kfed agrees strongly with the target clustering.  Note that by construction
all devices with data from the same set $G_i$ contain data from the same set of
Gaussian components. Further, devices with data from different sets $G_i$ have
no common Gaussian component.  Thus all cluster pairs within the same set $G_i$
are active cluster pairs and there are $\sqrt{k}{\sqrt{k} \choose 2}$ such
pairs. Moreover, any pair $(r, s)$ such that $r \in G_i$, $s \in G_j$ $i\ne j$
form an inactive cluster pair and there are ${k \choose 2} - \sqrt{k}{\sqrt{k}
\choose 2} = O(k^2)$ such pairs. These need only satisfy the weaker inactive
separation requirement.

\begin{figure}
    \centering
    \begin{subfigure}{0.22\textwidth}
    \includegraphics[width=\textwidth]{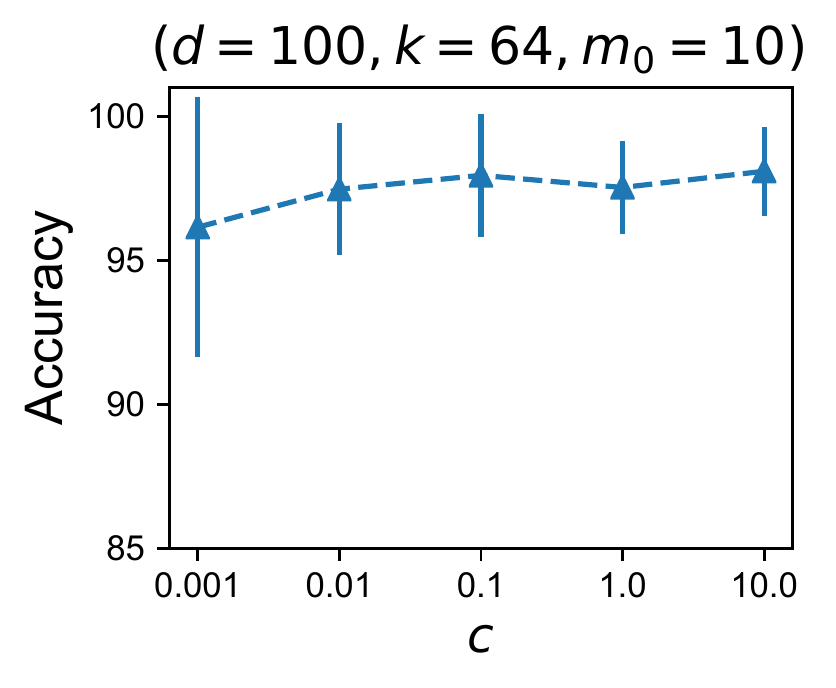}
    \end{subfigure}
    \hfill
    \begin{subfigure}{0.22\textwidth}
    \includegraphics[width=\textwidth]{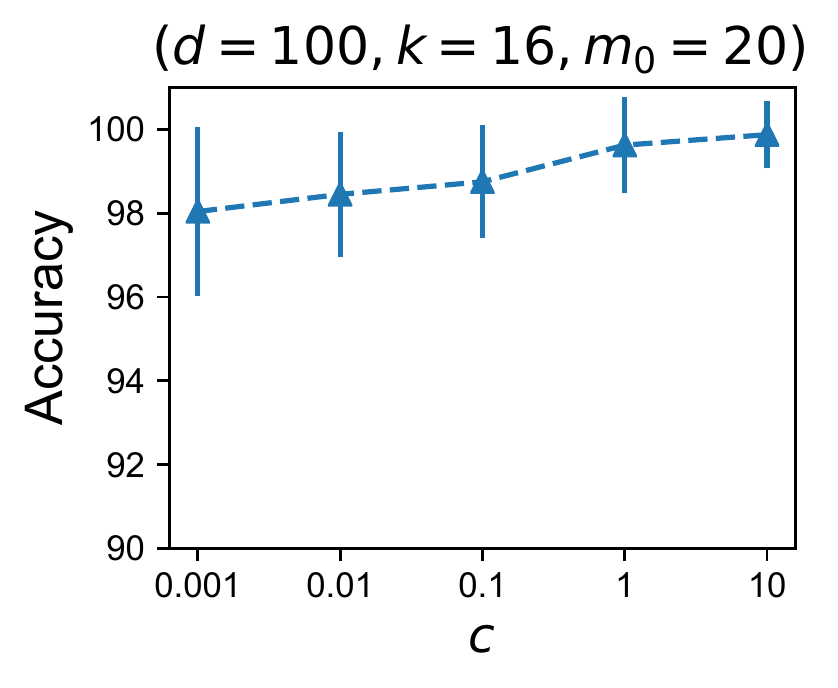}
    \end{subfigure}
    \caption{Impact of the separation constant $c$ on the clustering accuracy when
    clustering a mixture of Gaussians. Even for relatively small values of $c$,
    for the case of data generated from a mixture of Gaussians, \kfed can recover
    highly accurate clustering with decreasing variance across runs.}
    \label{fig:about-c}
\end{figure}

Note that while we prescribe $c \ge 100$ for our arguments to hold,
Figure~\ref{fig:about-c} demonstrates that clustering can be recovered even in
settings where $c$ is much smaller.

\subsection{Empirical Evaluation on Real Data}

In this section, we empirically explore \kfed and the related analyses from
Section~\ref{sec:method}. First, we validate our theoretical results, showing
that clustering over structured (heterogeneous) partitions can improve
clustering performance relative to clustering over random, IID partitioned
data. Second, we explore the effect of one-shot clustering relative to more
communication-intensive baselines. Finally, we investigate practical
applications of one-shot clustering in terms of client sampling and
personalized federated learning.

\subsubsection{Properties of \kfed}

\begin{figure}
    \centering
    \begin{subfigure}{0.23\textwidth}
    \includegraphics[width=\textwidth,trim=17 10 15 0 , clip]{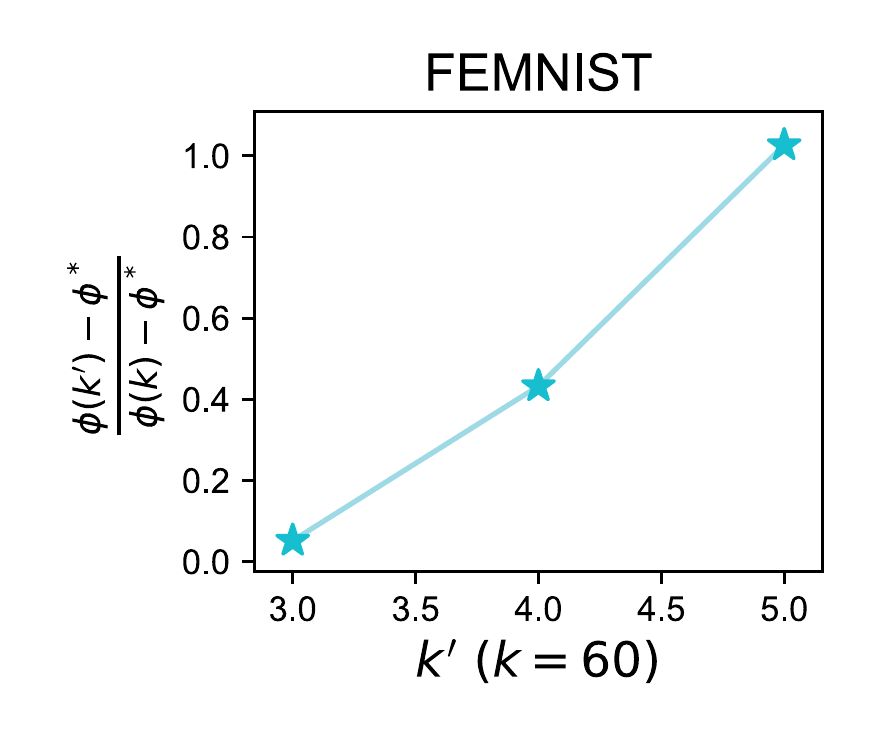}
    \end{subfigure}
    \hfill
    \begin{subfigure}{0.23\textwidth}
    \includegraphics[width=\textwidth, trim=17 10 15 0 , clip]{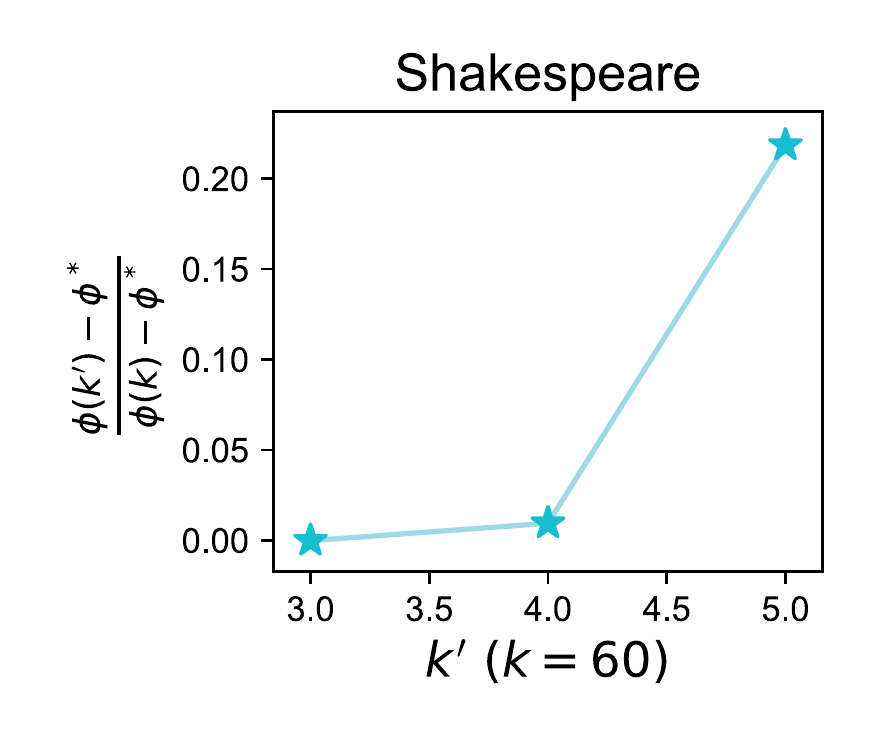}
    \end{subfigure}
    \caption{The $k$-means cost under structured partitions ($\phi(k')$) is
    closer to the cost of oracle clustering ($\phi^*$) than that under random
    partitions ($\phi(k)$). As heterogeneity increases ($k'$ decreases), the
    benefits of structured partitions are becoming more significant, with
    $\phi(k')-\phi^* \ll \phi(k)-\phi^*$.}
    \label{fig:cost_ratio}
\end{figure}

\textbf{Benefits of Heterogeneity (Def.~\ref{def:heterogeneity}).} We compare
the performance of \kfed on two different partitions of data among devices: (i)
one with \textit{IID random} partitions, and (ii) another with
\textit{structured} partitions.  To generate the underlying structured partition
for this experiment we use the following heuristic. First, we cluster all the data into $k$
clusters for a range of values of $k$. For each $k$, we take the clustering we
have as the target clustering $\mathcal{T}$, and construct the data matrix $A$ and
the matrix of centers $C$.  Finally,  for each pair of cluster means $\mu_r,
\mu_s$, we compute the quantity $\frac{\norm{\mu_r -
\mu_s}}{2\sqrt{m_0}(\Delta_r + \Delta_s)}$, the ratio of the actual separation
of the cluster mean to the required active separation. We pick a value of $k$
at which a large number of clusters are reasonably well separated (see
Appendix~\ref{app:datasets}, Figure~\ref{fig:mnist-sep}). We call this our
\textit{oracle clustering}. Now to generate the IID partition for (i), we
randomly distribute this data among $Z$ devices. To generate the structured
partition for (ii), we divide the data among $Z$ devices such that each device
receives only data from a random subset of no more than $k'$ clusters. For each
value of $k'$, we cluster the data for both cases over the devices using \kfed
and compute the $k$-means cost. Let $\phi^*$ denotes the $k$-means cost of the
original oracle clustering. Let $\phi(k')$ denote the $k$-means cost when $k'$
clusters are assigned to each device.  Figure~\ref{fig:cost_ratio} presents the
relative cost ratio between the cost change in structured partitions
($\phi(k')-\phi^*$) and random partitions ($\phi(k)-\phi^*$). 

We perform this
experiment on the FEMNIST and Shakespeare datasets~\cite{caldas2018leaf} (see Appendix~\ref{app:datasets}
for details). It can be seen from the results plotted in
Figure~\ref{fig:cost_ratio} that clustering on structured splits achieves a
cost closer to that of the oracle partition compared to the cost achieved on
the IID random partition.  We note that the separation achieved in real datasets is
much smaller than required even with this careful construction
(Appendix~\ref{app:datasets}). Even still, our experiments demonstrate that heterogeneity can benefit federated clustering on common benchmarks.

\textbf{Communication-Efficiency.} One advantage of the proposed method is that
it requires only a single round of communication. Given this, it is natural to
wonder how the performance of \kfed would compare with other, more
communication-intensive clustering baselines. In particular, a common way to
solve $k$-means in distributed settings is to simply parallelize the cluster
assignment and cluster mean calculations at each step.  Here, we show that for
different partitions of the dataset with multiple values of $k'$, our one-shot
method $k$-FED is able to produce similar clustering outputs (in terms of the
$k$-means cost; lower is better) as naive distributed $k$-means, which requires multiple
communication rounds. Here we use the same oracle clustering as the previous
experiment to construct our device data.

\begin{figure}[h!]
    \centering
    \begin{subfigure}{0.23\textwidth}
    \includegraphics[width=\textwidth]{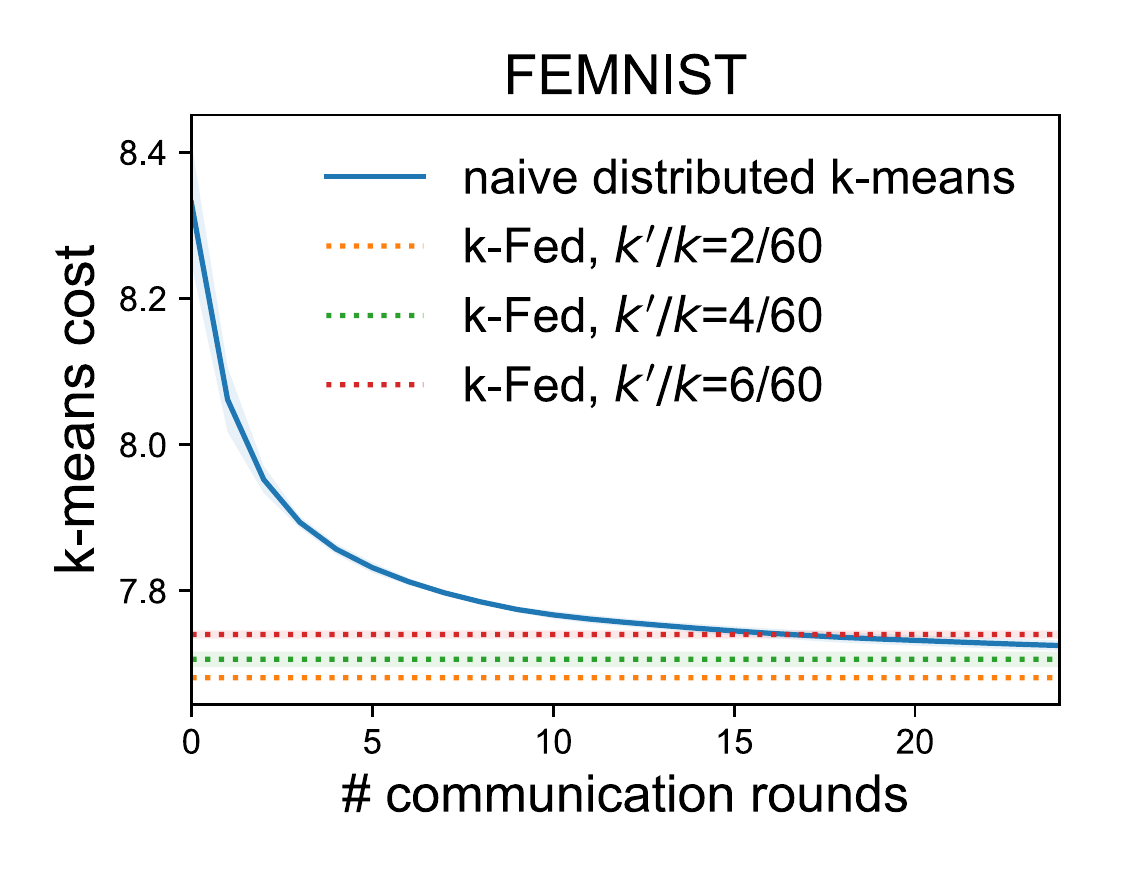}
    \end{subfigure}
    \hfill
    \begin{subfigure}{0.23\textwidth}
    \includegraphics[width=\textwidth]{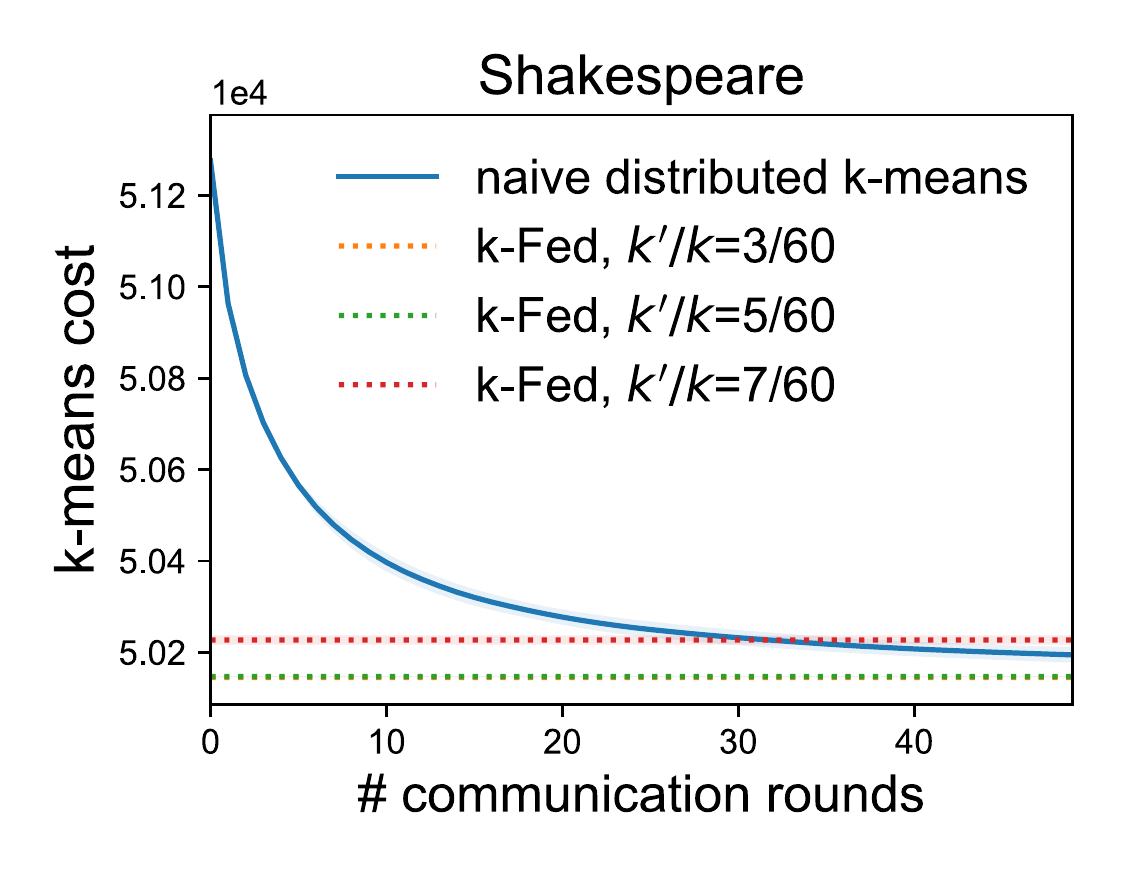}
    \end{subfigure}
    \caption{$k$-FED (using just one communication round) is able can provide
    similar clustering quality as naive distributed $k$-means.}
  \label{fig:communication}
\end{figure}\vspace*{-4mm}

\subsubsection{Applications of $k$-Fed}

\textbf{Personalized FL.} Compared with fitting a single global model to data
across all device, jointly learning personalized (separate but related) models
can boost the effective sample size while adapting to the 
heterogeneity in federated networks \citep[e.g.,][]{smith2017federated,mansour2020three}.

\citet{ghosh2020} recently proposed an algorithm to learn models over federated
networks where devices are partitioned into clusters when the clustering
information is unavailable. Consider a supervised learning problem that each
cluster of devices want to solve and assume the number of clusters $k$ is
known. Their method, the Iterative Federated Clustering Algorithm (IFCA), in
its first step initializes $k$ models $(m_1, \dots, m_k)$, one for each
cluster.   At the start of each round, all $k$ models are sent to the
devices. Each device picks the model that minimizes a loss function on its
locally available data. The device can be configured to now either compute and
transmit the gradient of the loss function of this model or it can perform a
few model updates locally and send the updated model to the central server. As
the last step of the round, for each model $m_i$ $i\in[k]$, all the devices
that picked this model are identified. All these devices are assigned cluster id
$i$. Model $m_i$ then is updated by either model averaging or gradient
averaging using the information sent by devices in cluster $i$. 

\begin{table}
  \caption{Test accuracy of rotated MNIST on three methods. Training
  personalized models based on the clustering information output by \kfed
  achieves the same performance of IFCA, without the high computation and
  communication overhead of IFCA when $k'=1$. For $k'=2$, the performance of
  \kfed degrades much less when compared to that of IFCA.}
  \vspace{1em}
  \centering
  \label{table:personalization}
  \begin{tabular}{l ccc} 
     \toprule[\heavyrulewidth]
         & \textbf{Global} & \textbf{IFCA} & \textbf{$k$-FED}  \\
        \midrule
        100 devices ($k'=1$) & 95.0  & 98.0 & 98.0 \\
        200 devices ($k'=1$) & 94.5  & 97.2 & 97.8 \\
        \midrule
        100 devices ($k'=2$) & 95.3 & 95.6 & 97.1\\
        200 devices ($k'=2$)  & 94.5  & 95.1 & 96.4\\
    \bottomrule[\heavyrulewidth]
  \end{tabular}
\end{table}
We instantiate IFCA on the problem of learning personalized models for
clusters. As in~\cite{ghosh2020}, we use the MNIST dataset for this experiment.
We construct $k=4$ clusters by $0, 90, 180$ and $270$ degree rotations and
distribute them among devices. Note that in the setup for IFCA, each device
only contains data from a single cluster (since we are clustering devices and
not individual data points). Thus we set $k'=1$ and compare IFCA with a simple
\kfed based method: We first perform one-shot clustering to obtain an initial
clustering and then we use FedAvg~\cite{mcmahan2017communication} to learn one
model per cluster. As a baseline, we also learn a single global model and
include it for comparison. As can be seen from the test accuracies in
Table~\ref{table:personalization} ($k'=1$), \kfed is competitive with IFCA.
Moreover, \kfed has the additional advantage that once the cluster identities
have been assigned, we only need to transmit one model instead of the $k$ models that are transmitted with IFCA.

Since \kfed clusters data, the \kfed + FedAvg approach can also handle cases
where there are data from multiple clusters on the same device.
Table~\ref{table:personalization} ($k'=\sqrt{k}=2$) shows the test accuracy on
such a partition. Here we observe the performance of IFCA degrade when compared
to \kfed.

\begin{figure}[h!]
    \centering
    \begin{subfigure}{0.22\textwidth}
    \includegraphics[width=\textwidth]{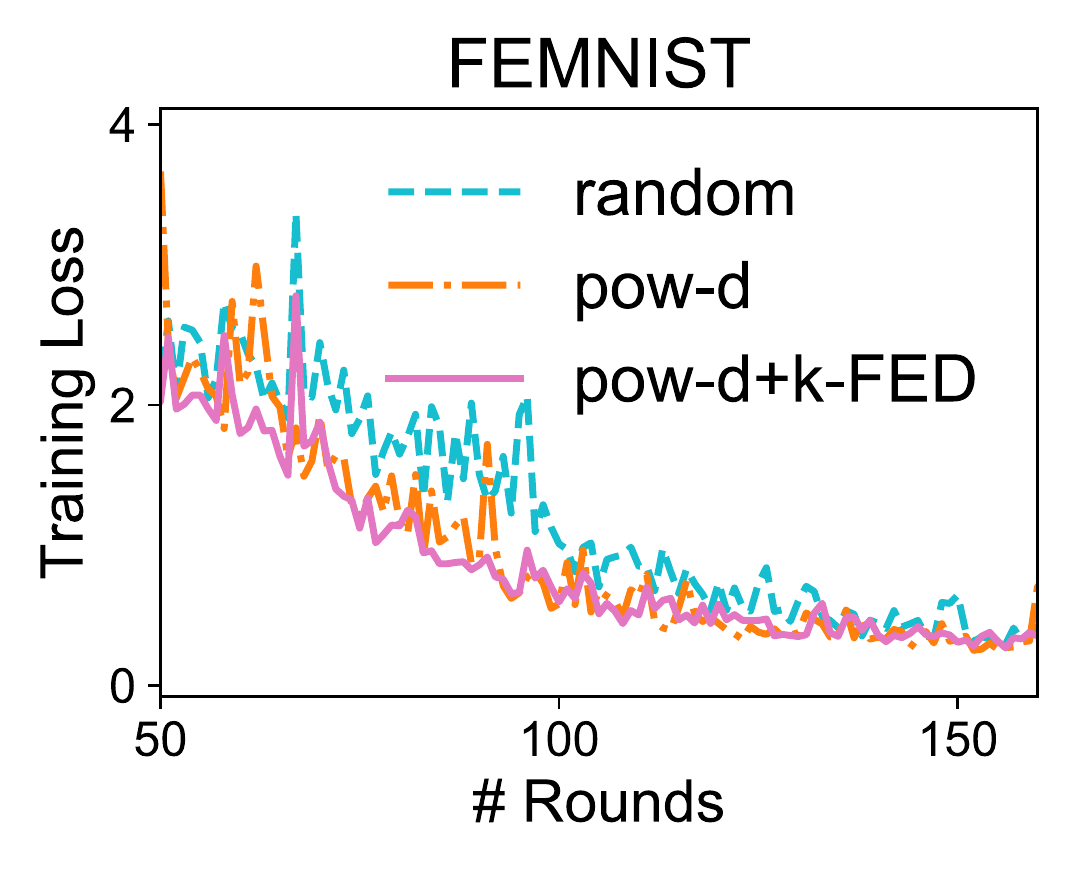}
    \end{subfigure}
    \hfill
    \begin{subfigure}{0.22\textwidth}
    \includegraphics[width=\textwidth]{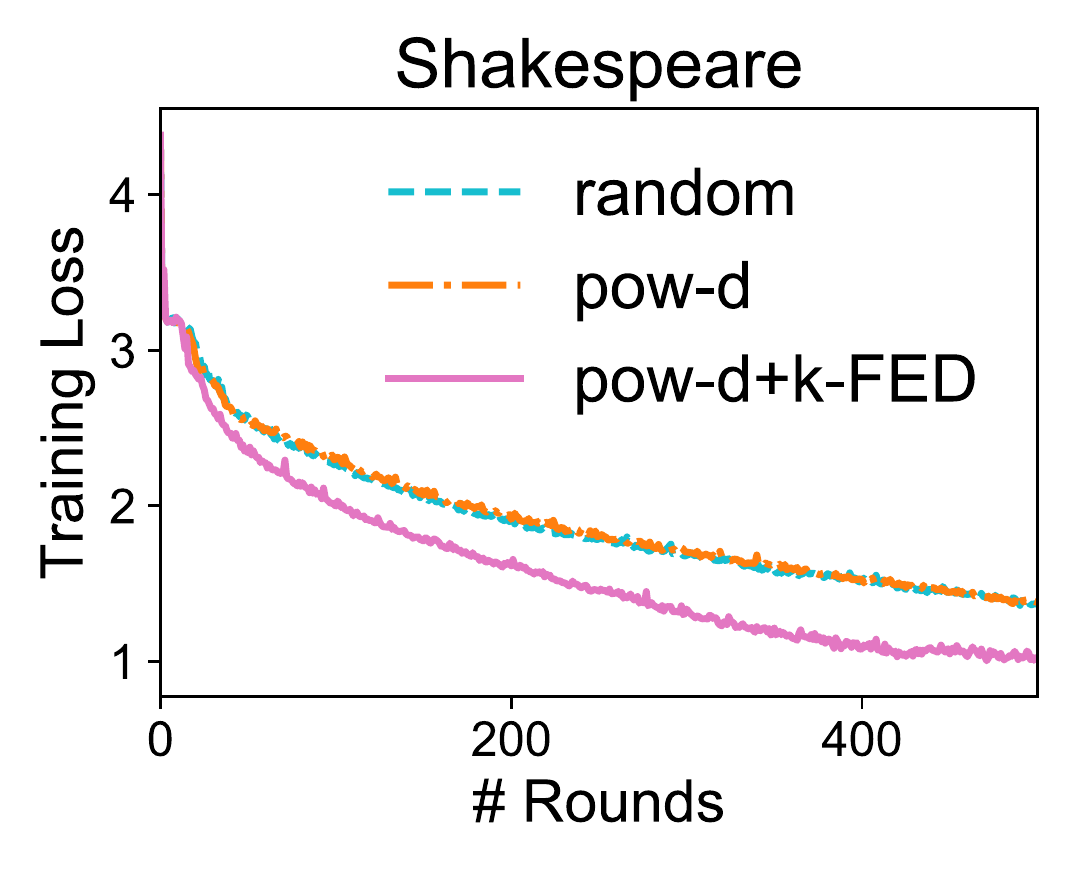}
    \end{subfigure}
    \caption{Additional clustering information provided by $k$-FED can help
    achieve faster convergence than recent client selection techniques
    pow-d~\cite{cho2020client}.}
    \label{fig:client_selection}
\end{figure}

\textbf{Client Selection.} Finally, we demonstrate that the clustering information produced
by $k$-FED is a useful prior for client selection
applications~\cite{cho2020client}. In practice, cross-device federated optimization
algorithms need to tolerate partial device participation~\cite{kairouz2019advances}. Intuitively, incorporating
information from  `representative' devices at each communication round may
speed up the convergence of learning tasks over federated networks as opposed to
randomly sampling devices. When randomly sampling, similar and potentially
redundant clients can be selected. A recent device
selection method proposes to additionally select the devices with large
training losses among those randomly-selected subset of
devices~\cite{cho2020client} to help with convergence speed. We combine $k$-FED
with this approach by further filtering out the devices coming from the same
clusters. Note that $k$-FED does not add significant additional overhead to the
baseline algorithm as it only requires running one-shot clustering before
training. The results are shown in Figure~\ref{fig:client_selection}. We see
that leveraging the underlying structure learnt by $k$-FED can boost
convergence on these realistic federated benchmarks.

Similar to~\citet{cho2020client}, we also observe that for the experiments in Figure~\ref{fig:client_selection}, the variance of test
performance across all devices has been reduced using client selection
strategies favoring more informative (potentially more underrepresented)
clients compared with that of random selection. 
For instance, on FEMNIST, the variance of final test accuracies is reduced by 35\% when using k-fed combined with pow-d instead of random selection.
This may be useful
in scenarios where we wish to impose notions fairness for federated learning~\cite{mohri2019agnostic, li2019fair}.

\section{Conclusion}
In this work, we provide an example of how heterogeneity in federated networks
can be beneficial, by rigorously analyzing the effects of heterogeneity on a
simple, one-shot variant of Lloyd's algorithm for distributed clustering. Our
proposed method, \kfed, addresses common practical concerns  in  federated
settings,  such  as  high  communication costs,  stragglers,  and  device
failures.  We believe that other, specific notions of heterogeneity---together
with careful analyses---may provide benefits for a plethora of other problems
in federated learning, which is an interesting direction of future work.

\section{Acknowledgements} 
This work  was supported in part by the
National Science Foundation Grant IIS1838017, a Google Faculty Award, a Facebook Faculty Award, and the CONIX Research Center. Any opinions, findings, and conclusions or
recommendations expressed in this material are those of the author(s) and do not necessarily reflect
the National Science Foundation or any other funding agency.

\newpage
\bibliography{references}
\bibliographystyle{icml2021}

\clearpage\newpage
\onecolumn
\appendix
\section{Proofs}
\label{app:proofs}

\subsection{Proving Theorem ~\ref{th:gen} (Main Theorem)}
Before we proceed to proving Theorem~\ref{th:gen}, we first establish a few
preliminary results. Let $\mathcal{T} = (T_1, \dots, T_k)$ be our target
clustering and let  $T^\devz_r$ be the subset of points of a cluster $T_r$ on
device $z$. For any point, $A^\devz_i$ on device $z$, let $c(A^\devz_i)$ denote
the index of the cluster it belongs to. That is, 
\[
  A^\devz_i \in T^\devz_{c(A^\devz_i)} \subseteq T_{c(A^\devz_i)}.
\] Also recall the definition of matrix $C$, the matrix of means. Here the
$i$-th row of $C$ contains the mean of the cluster which contains data points
$A_i$, i.e.  $C_i = \mu(T_{c(A_i)})$.  Our first lemma bounds how far the
`local' cluster mean $\mu(T^\devz_r)$ can deviate from $\mu(T_r)$.

\begin{lemma}[Lemma 5.2 in~\citet{kumarkannan01}] Let $T^\devz_r$ be a subset
  of $T_r$ on device $z$. Let $\mu(T^\devz_r)$ denote the mean of the points
  indexed by $T^\devz_r$.  Then,
\begin{align*}
  \ltwos{\mu(T^\devz_r) - \mu(T_r)} \le \frac{\op{A-C}}{\sqrt{\abs*{T^\devz_r}}}.
\end{align*}
\label{lemma:meanshift}
\end{lemma}
\begin{proof} Let $A^\devz$ be the sub-matrix of $A$ on device $z$ and let
  $\tilde{C}^\devz$ be the corresponding sub-matrix of our matrix of means $C$.
  Let $u$ be an indicator vector for points in $T^\devz_r$.  Observe that,
  \begin{align*}
    \ltwo{\ \abs*{T^\devz_r}(\mu(T^\devz_r) - \mu_r)\ } 
      &= \ltwos{(A^\devz - \tilde{C}^\devz)\cdot u}\\
      &\le \ops{A^\devz -\tilde{C}^\devz}\ltwo{u}\\
      &\le \ops{A-C}\sqrt{\abs{T^\devz_r}}.
  \end{align*} Here, for the last inequality, we note that
  $(A^\devz-\tilde{C}^\devz)$ contains a subset of rows of $(A-C)$, and
  therefore ${\ops{A^\devz - \tilde C ^\devz} \le \ops{A-C}}$.
\end{proof}
Now consider the local clustering problem on each device $z$. The device has a
data matrix $A^\devz$, whose rows are a subset of $A$. Let $T^\devz_1,
T^\devz_2, \dots, T^\devz_{k}$ be subsets of $T_1, T_2, \dots, T_k$ on this
device, such that no more than $k'$ of them are non-empty. Construct a matrix
${C^\devz}$, of the same dimensions as $A^\devz$ where for each row of
$A^\devz$, the corresponding row of $C^\devz$ contains the mean of the local
cluster the point belongs to. That is, the $i$-th row of $C^\devz$ contains
$\mu(T^\devz_{c(A^\devz_i)})$. Using this next lemma, we bound the operator
norm of the matrix $(A^\devz-C^\devz)$, in terms of $(A-C)$.
\begin{lemma} \label{lemma:normchange}Let $T^\devz_1, T^\devz_2, \dots
  T^\devz_k$ be subsets of target cluster that reside on a device such that
  $k'$ of them are non-empty. Let $A^\devz$ be the corresponding $n^\devz\times
  d$ data matrix.  Let $C^\devz$ be the corresponding matrix of means; that is
  each row $C^\devz_i = \mu(T^z_{c(A^z_i)})$. Then,
  \[
    \ops{A^\devz-C^\devz} \le 2\sqrt{k'}\ops{A-C} \, .
  \]
\end{lemma}
\begin{proof} Let $\tilde{C}^\devz$ be an $n^\devz\times d$ matrix where
  $\tilde{C}^\devz_i = \mu(T_{c(A^\devz_i)})$. First, consider a unit vector
  $u$ along the top singular direction and observe that:
  \begin{align*}
    \ops{\tilde C^\devz - C^\devz}^2 
      &=\sum_{r=1}^{k}\abs*{T^\devz_r}\Big(\big(\mu(T^\devz_r) -\mu(T_r)\big)
        \cdot u\Big)^2\\
      &\le \sum_{r=1}^{k}\abs*{T^\devz_r}\ltwo{\mu(T^\devz_r) -\mu(T_r)}^2\\
      &\le_{(a)} k'\op{A-C}^2.
  \end{align*} Here for inequality $(a)$ we invoke Lemma~\ref{lemma:meanshift}.
  Also, noting that $\ops{A^\devz-\tilde{C}^\devz} \le \ops{A-C}$, we get,
  \begin{align*}
    \ops{A^\devz-C^\devz} &\le \ops{A^\devz-\tilde{C}^\devz} 
    + \ops{\tilde{C}^\devz-C^\devz}\\
    &\le (1+ \sqrt{k'})\ops{A-C} \le 2\sqrt{k'}\ops{A-C}.
  \end{align*}
\end{proof} 
We prove Theorem~\ref{th:gen} in four parts:
\begin{enumerate}
  \item In the first step we show that satisfying the active separation
    condition is sufficient to satisfy the Awasthi-Sheffet separation condition
    required for Lemma~\ref{lemma:awasthi} (Lemma~\ref{lemma:centersep2}).

  \item Next we use Lemma~\ref{lemma:centersep2} to show that the first step of
    \kfed (\algoA) will find local centers $\theta^\devz_r$ that are close to
    true centers $\mu(T^\devz_r)$ on device $z$. We state and prove this in
    Lemma~\ref{lemma:localbound}.

  \item In next step, we show that the process of picking $k$ initial centers
    in steps 2-6 of \kfed picks exactly one local cluster center
    $\theta^\devz_r$ for each cluster $r$.  That is, we pick $k$ local centers
    one corresponding to each target cluster. (Lemma~\ref{lemma:center_init})

  \item Finally, we argue that with this initialization, the clustering of
    local cluster centers produced $(\tau_1, \dots, \tau_k)$ has the property
    that, all local cluster centers corresponding the to the same cluster (say
    $T_r$) will be in the same set (say $\tau_r$). Moreover, no local cluster
    center corresponding to any $T_s$, $s\ne r$ will be in $\tau_r$. As we
    argue later, this is sufficient for the induced clustering produced by
    $(\tau_1,\dots, \tau_k)$ to agree with our target clustering $\mathcal{T} =
    (T_1, T_2, \dots)$ up to permutation of labels and missclassifications
    incurred at the local clustering stage.
\end{enumerate} 

\begin{lemma}\label{lemma:centersep2} Let $(T_r, T_s)$ be cluster pairs such
  that, $\ltwo{\mu_r - \mu_s} \ge 2c\sqrt{m_0}(\Delta_{r} + \Delta_s).$ Let
  $T^z_r \subseteq T_r$ and $T^z_s \subseteq T_s$ be large subsets on device
  $z$. Then,
  \begin{align*}
    \ltwos{\mu^\devz_r - \mu^\devz_s} \ge c\sqrt{k'}
    \left(\frac{\ops{A^\devz - C^\devz}}{\sqrt{n^\devz_r}} + \frac{\ops{A^\devz
    - C^\devz}}{\sqrt{n^\devz_s}}\right).
  \end{align*}
\end{lemma}

\begin{proof} (Lemma~\ref{lemma:centersep2})
  Using the triangle inequality, we have 
  \begin{align*}
    \ltwos{\mu^\devz_r - \mu^\devz_s}
      &\ge\ltwos{\mu_r - \mu_s}-\ltwos{\mu^\devz_r-\mu_r}
      -\ltwos{\mu_s-\mu^\devz_s}\\
      & \ge 2c\sqrt{m_0}(\Delta_r + \Delta_s)
      - \frac{\ops{A-C}}{\sqrt{n^\devz_r}}
        -\frac{\ops{A-C}}{\sqrt{n^\devz_s}}\numberthis\label{eq:proof:loc1}
  \end{align*} using the active separation assumption. Now, expanding the terms
  can write the left hand side as
  \begin{align*}
    \ltwos{\mu^\devz_r - \mu^\devz_s} &\ge
      2c\sqrt{m_0}\left( k' \frac{\ops{A-C}}{\sqrt{n_r}} +  k'
      \frac{\ops{A-C}}{\sqrt{n_s}}\right)
      -
      \frac{\ops{A-C}}{\sqrt{n^\devz_r}}-\frac{\ops{A-C}}{\sqrt{n^\devz_s}}\\
    &\ge\underbrace{\left(2\sqrt{\frac{m_0 n^\devz_r}{n_r}} - \frac{1}{ck'}\right)
    ck'\frac{\ops{A-C}}{\sqrt{n^\devz_r}}}_{(i)}
    + \underbrace{\left(2\sqrt{\frac{m_0 n^\devz_s}{n_s}} - \frac{1}{ck'}\right)
    ck'\frac{\ops{A-C}}{\sqrt{n^\devz_s}}}_{(ii)}.
  \end{align*} We first only consider the term $(i)$. According to
  Lemma~\ref{lemma:normchange}, $\ops{A-C} \ge \frac{1}{2\sqrt{k'}}\ops{A^\devz
  - C^\devz}$.  Using this we can bound $(i)$ as
  \[
    \left(2\sqrt{\frac{m_0 n^\devz_r}{n_r}} - \frac{1}{ck'}\right)
    ck'\frac{\ops{A-C}}{\sqrt{n^\devz_r}} 
    \ge
    \left(2\sqrt{\frac{m_0 n^\devz_r}{n_r}} - \frac{1}{ck'}\right)
    c\sqrt{k'}\frac{\ops{A^\devz-C^\devz}}{\sqrt{n^\devz_r}}\, .
  \] Now recall that for large cluster subsets $n^\devz_r \ge
  \frac{1}{m_0}n_r$ and thus $2\sqrt{\frac{m_0 n_r^\devz}{n_r}}  -
  \frac{1}{ck'} \ge 2 - \frac{1}{ck'} \ge 1$. This means that we can bound term
  $(i)$ as,
  \[
    \left(2\sqrt{\frac{m_0 n^\devz_r}{n_r}} - \frac{1}{ck'}\right)
    ck'\frac{\ops{A-C}}{\sqrt{n^\devz_r}} 
    \ge
    c\sqrt{k'}\frac{\ops{A^\devz-C^\devz}}{\sqrt{n^\devz_r}} \, .
  \] We get a symmetric expression for term $(ii)$ as well.  Using
  this in equation~\ref{eq:proof:loc1}, we get the desired result:
  \begin{align*}
    \ltwos{\mu^\devz_r - \mu^\devz_s}
    \ge 
    c\sqrt{k'}\biggl(\frac{\ops{A^\devz-C^\devz}}{\sqrt{n^\devz_r}} + 
    \frac{\ops{A^\devz-C^\devz}}{\sqrt{n^\devz_s}}\biggr) \, .
  \end{align*}
\end{proof}
Since \algoA is run locally on each device, it is unaffected by the inactive
separation condition, as by definition, subsets of only active cluster pairs
exist on each device. This implies that \algoA solves the local clustering
problem successfully. Specifically on device $z$ containing data from some
cluster $T_r$, $\theta^z_r$ is not too far from $\mu(T^\devz_r)$.  Showing this
result is our second step and we state this formally in
Lemma~\ref{lemma:localbound} below.

\begin{lemma}\label{lemma:localbound} Let $(T_1^\devz, \dots, T_k^\devz)$ be
  the subsets of $(T_1, \dots, T_k)$ on some device $z$ such that no more than
  $k'$ of them are non-empty. Moreover, assume all non-empty subsets are large,
  i.e.  $\abs*{T^\devz_r} \ge \frac{1}{m_0} \abs{T_r}$. Finally, assume that
  the active separation requirement is satisfied for all active cluster pairs
  on $z$. Then, on termination of~\algoA, for each non-empty $T^\devz_r$, we
  have
  \[
    \ltwos{\theta^\devz_r - \mu(T^\devz_r)} \le
    \frac{25}{c}\frac{\ops{A^\devz-C^\devz}}{\sqrt{n^z_r}}
    \le 
    \frac{50\sqrt{k'}}{c}\frac{\ops{A-C}}{\sqrt{n^z_r}} \, ,
  \] and,
  \[
    \ltwos{\theta^\devz_r - \mu(T_r)} \le
    2\sqrt{m_0k'}\frac{\ops{A-C}}{\sqrt{n_r}} \le
    2\sqrt{m_0}\lambda \, .
  \] 
\end{lemma}
\begin{proof} First note that the local clustering problem with data matrix
  $A^\devz$ and matrix of centers $C^\devz$ satisfies the requirements of
  Lemma~\ref{lemma:awasthi}. Thus it follows that,
  \[
    \ltwos{\theta^\devz_r - \mu(T^\devz_r)} \le
    \frac{25}{c}\frac{\ops{A^\devz-C^\devz}}{\sqrt{n^z_r}} \, .
  \] Now applying Lemma~\ref{lemma:normchange} gives us the first statement.
  
  To prove the second statement, we start off with the triangle inequality:
  \begin{align*}
    \ltwos{\theta^\devz_r - \mu(T_r)} 
    &\le \ltwos{\theta^\devz_r - 
    \mu(T^\devz_r)}  + \ltwos{\mu(T^\devz_r) - \mu(T_r)}\\
    &\le \frac{25}{c}\frac{\ops{A^\devz- C^\devz}}{\sqrt{n_r^z}} 
    + \frac{\ops{A- C}}{\sqrt{n_r^z}} \, .
  \end{align*} Here for the last inequality we used
  Lemma~\ref{lemma:meanshift}. Now applying Lemma~\ref{lemma:normchange} and
  taking take $c \ge 100$, we get
  \begin{align*}
    \ltwos{\theta^\devz_r - \mu(T_r)} 
    &\le \frac{50\sqrt{k'}}{c}\frac{\ops{A- C}}{\sqrt{n_r^z}}
    + \frac{\ops{A- C}}{\sqrt{n_r^z}}\\
    &\le \left(\frac{50}{c} + \frac{1}{\sqrt{k'}}\right)
    \sqrt{k'}\frac{\ops{A- C}}{\sqrt{n_r^z}}\\
    &\le 2\sqrt{k'}\frac{\ops{A-C}}{\sqrt{n^z_r}}
    \le 2\sqrt{m_0 k'}\frac{\ops{A-C}}{\sqrt{n_r}}\\
    &\le 2\sqrt{m_0}\lambda \, .
  \end{align*} 
\end{proof}

This means that for a fixed $r$, all the $\theta^\devz_r$ received at the
central server from devices $z \in [Z]$ are `close' to $\mu(T_r)$. 

The next step is to show that in the $k$ initial centers \kfed picks in steps
2-6, there is exactly one corresponding to each target cluster $T_i$. We will
show later that this is sufficient for the final step of the algorithm to
correctly assign local cluster centers to the correct partition.
\begin{lemma}\label{lemma:center_init} Let $\mathcal{T} = (T_1, \dots, T_k)$ be
  our target clustering.  Assume all active cluster pairs and inactive cluster
  pairs satisfy their separation requirements. Further let $n_{\min} \ge
  \frac{4}{c^2k'}n_{\max}$. Then at the end of step 6 of \kfed, for every
  target cluster $T_r$, there exists an $\theta^\devz_s \in M$ such that
  $\theta^\devz_s = \mu(T^\devz_s)$ for some $z \in [Z]$.
\end{lemma}

Before we proceed to proving this lemma, we state and prove a lower bound on
how close a local cluster center $\theta_r^\devz$ can be to some cluster mean
$\mu(T_s)$ for $s \ne r$:

\begin{lemma}\label{lemma:missclass_lb} Let $\theta^\dev{z}_r :=
  \mu(T^\devz_r)$.  The for any $s \ne r$, $z' \in [Z]$,
  \[
    \ltwos{\theta^\devz_r - \theta^\dev{z'}_s}
    \ge 6 \sqrt{m_0}\lambda \, .
  \]
\end{lemma}

\begin{proof} First, from the triangle inequality note that,
  \[
    \ltwos{\theta^\devz_r -\theta^\dev{z'}_s} 
    \ge \ltwos{\mu_r - \mu_s} - \ltwos{\mu_r -\theta^\devz_r}
    - \ltwos{\mu_s
    -\theta^\dev{z'}_s} \, .
  \] Using Lemma~\ref{lemma:localbound} and our inactive separation
  assumption we bound the right hand side further as,
  \begin{align*}
    \ltwos{\mu_r - \mu_s} - \ltwos{\mu_r -\theta^\devz_r}
    - \ltwos{\mu_s -\theta^\dev{z'}_s}
    &\ge10\sqrt{m_0k'}\frac{\ops{A-C}}{\sqrt{n_{\min}}} 
    -4\sqrt{m_0 k'}\frac{\ops{A-C}}{\sqrt{n_r}}\\
    &\ge
    6\sqrt{m_0k'}\frac{\ops{A-C}}{\sqrt{n_{\min}}} 
    \ge 6\sqrt{m_0}\lambda,
   \end{align*} as desired.
\end{proof}

\begin{proof}(Lemma~\ref{lemma:center_init}) Let $M_t$ denote the set $M$ in
  step 2-6 of \kfed, after picking the first $t$ points $(1 \le t \le k)$.
  Let us denote the point \kfed selects in iteration $t$ as $\theta_t$. That is,
  \[
    \theta_t \leftarrow \argmax_{z \in [Z], i \in
    [k]}d_{M_{t-1}}(\theta^\devz_i) \, .
  \]
  We will show that the set $M_t$ contains $t$ points corresponding to $t$
  different target clusters at every iteration $t$. This invariant holds
  trivially at $t=1$. Assume the statement first became false at some $1 < t'
  \le k$.  Let the point $\theta_{t'}$ correspond to a local cluster mean from
  cluster $T_r$.  Then there must exist some $1 \le t'' < t'$ such that
  $\theta_{t''}$ also correspond to a local cluster mean from $T_r$.  Further,
  there must exist some cluster $s\ne r$ such that $\theta^\devz_s \not\in
  M_{t'}$ for any $z \in [Z]$. 
  
  Now by definition of $d_{M_{t-1}}(\theta_{t'})$, we have
  \begin{align*}
    d_{M_{t-1}}(\theta_{t'}) 
    &= \min_{\theta \in M_{t-1}}\ltwos{\theta_{t'} -
    \theta}\\
    &\le \ltwos{\theta_{t'} - \theta_{t''}}\\
    &\le_{(a)} \ltwos{\theta_{t'} - \mu(T_r)} + \ltwos{\mu(T_r) -
    \theta_{t''}}\\
    &\le_{(b)}
    4\sqrt{m_0k'}\frac{\ops{A-C}}{\sqrt{n_r}}
    \le 4\sqrt{m_0}\lambda \, . \label{eq:loc:1}\numberthis
  \end{align*} Here inequality (a) follows from the triangle inequality and (b)
  follows from Lemma~\ref{lemma:localbound}.

  Now consider $\theta^\devz_s$ for any $z$. Since no other local cluster
  center from $T_s$ is contained in $M_t$, from
  Lemma~\ref{lemma:missclass_lb} we conclude that for every $\theta \in
  M_{t-1}$,
  \begin{align*}
    \ltwos{\theta^\devz_s -\theta} \ge 6\sqrt{m_0}\lambda \, .
  \end{align*}
  But this means that $d_{M_{t-1}}(\theta_{t'}) \le
  4\sqrt{m_0}\lambda \le 6\sqrt{m_0}\lambda \le d_{M_{t-1}}(\theta^\devz_s)$
  leading to a contradiction based on the definition of $\theta_{t'}$. This
  completes our argument.
\end{proof}

Now we are ready to prove our main Theorem~\ref{th:gen}.

\begin{proof} From Lemma~\ref{lemma:center_init}, we know that the set $M$ at
  the end of step 6 of \kfed contains exactly one center corresponding to each
  target clustering.  Let the local cluster center $\tilde{\theta}_r \in M$
  correspond to the cluster $T_r$. Observe that for any $z \in [Z]$,
  \begin{align*}
    \ltwos{\theta^z_r - \tilde{\theta}_r} 
    &\le \ltwos{\theta^z_r - \mu_r} + \ltwos{\mu_r - \tilde{\theta}_r}\\
    &\le 4\sqrt{m_0}\lambda,
  \end{align*} using Lemma~\ref{lemma:missclass_lb}. Further, for any $s \ne
  r$,
  \begin{align*}
    \ltwos{\theta^z_s - \tilde{\theta}_r} 
    &\ge 6\sqrt{m_0}\lambda.
  \end{align*} This means that for every $r$ and $z\in[Z]$, $\theta^z_r$ is
  closer to the corresponding initial center $\tilde{\theta}_r$ than to any
  other initial center $\tilde{\theta}_s$, $s\ne r$. Let $\tau_r$ be the set of
  local cluster centers assigned to $\tilde{\theta_r}$. Then it can be seen
  that $\tau_r$ only contains local cluster centers $\theta^\devz_r$ for all
  devices $z$, i.e. $\tau_r$ contains all the device cluster centers
  corresponding to target cluster $T_r$.

  Now consider the definition of \kfed induced clustering
  (Definition~\ref{def:induced-clustering}), where we define
  \[
    T'_r = \set{i: A^\devz_i \in U_s^\devz\text{ and }
    \theta^\devz_s \in \tau_r}.
  \] In this case, we know that only local cluster centers corresponding to
  cluster $T_r$ is contained in $\tau_r$. Thus our induced cluster $T'_r$
  becomes,
  \[
    T'_r = \set{i: A^\devz_i \in U_r^\devz}.
  \]  Now from Lemma~\ref{lemma:awasthi} we know that on each device the sets
  $(U^\devz_1, \dots, U^\devz_{k'})$ and $(T^\devz_1, \dots, T^\devz_{k'})$
  only differ on at most $O(\frac{1}{c^2})n^\devz$. Summing this error over all
  devices $z$, we see that our induced clustering $(T'_1, \dots, T'_k)$ and the
  target clustering $(T_1, \dots, T_k)$ differ only on $O(\frac{1}{c^2})n$
  points. Finally, if all the local points satisfy their respective proximity
  condition (Definition~\ref{def:proxcondition}), then no points are
  missclassified. This concludes our proof.
\end{proof}

\subsection{Running Time of $\kfed$ and Handling New Devices}

We now analyze the running time of $\kfed$ steps 2-8. Since step 1 is running
\algoA on individual devices, we do not include the running time of this step
as part of our analysis. Note that with the separation assumptions in place,
\algoA will converge in polynomial time. However, as observed in practise,
Lloyd like methods typically only take a few iterations to terminate.

\begin{theorem} Steps 2-8 of \kfed takes $O(Zk'\cdot k^2)$
  pairwise distance computations to terminate. Further, after the set $M$ in
  step 6 has been computed, new local cluster centers $\Theta^\devz$ from a yet
  unseen device $z$ can be correctly assigned in $O(k'\cdot k)$ distance
computations.
\end{theorem}
\begin{proof} (Theorem~\ref{th:runtime}) The proof of the first part follows
  from a simple step by step analysis. Step 1 can be performed in $O(1)$. Step
  2-6 executes exactly $k$ times. At each iteration $t$, $(1 \le t\le k)$, we
  compute the distance of all device cluster centers, of which there are most
  $Zk'$, to the points in $M_{t-1}$.  Thus at iteration $t$, this can be
  implemented with $Zk'\cdot t$ distance computations.  Summing over all $t$,
  we see that steps 2-6 can run in $O(Zk' \cdot k^2)$ distance computations.
  Finally, step 7 requires us to assign all the $Zk'$ device cluster centers to
  one of the $k$ initial points in $M$.  This can be implemented in $O(Zk'\cdot
  k)$ distance computations.  Thus the overall complexity in terms of pairwise
  distance computations is $O(Zk'\cdot k^2)$.

  The second part of the statement follows from noting that for each
  $\theta^\devz_r \in \Theta^\devz$, the nearest point in set $M$ must be the
  initial center $\tilde{\theta}_r$ we picked as was demonstrated in the proof
  of Theorem~\ref{th:gen}. Thus every $\theta^\devz_r \in \Theta^\devz$ is
  assigned to the correct partition $\tau_r$ as required.
\end{proof}

\subsection{Separating Data from Mixture of Gaussian}

We now prove Theorem~\ref{th:mixgauss}. Recall that we are working in the
setting where $k' \le \sqrt{k}$. Our proof builds on results from
Lemma~6.3, \citet{kumarkannan01}.
\begin{proof} First consider an active cluster pair $r, s$. Based on our
  separation requirement, we have:
  \begin{align*}
    \ltwos{\mu_r - \mu_s} 
    &\ge \frac{2c\sqrt{km_0}\sigma_{\max}}{\sqrt{w_{\min}}}
    \text{polylog}\left(\frac{d}{w_{\min}}\right)\\
    &\ge 2c\sqrt{km_0}\frac{\sigma_{\max}\sqrt{n}}{\sqrt{w_{\min}n}}
    \text{polylog}\left(\frac{d}{w_{\min}}\right) \, .
  \end{align*} We further simplify the right hand to get,
  \[
    \ltwos{\mu_r - \mu_s} 
    \ge c\sqrt{km_0}\sigma_{\max}\sqrt{n}\bigl(\frac{1}{\sqrt{w_r n}}
    + \frac{1}{\sqrt{w_s n}}\bigr)
    \text{polylog}\left(\frac{d}{w_{\min}}\right) \, .
  \] Now note the number of points from each component $F_r$ is
  very close to $w_r n_r$ with very high probability. Here $w_r$ is the mixing
  weight of component $r$ and $n_r$ is the number of data points. Using this,
  with high probability we have
  \[
    \ltwos{\mu_r - \mu_s} 
    \ge c\sqrt{km_0}\sigma_{\max}\sqrt{n}\bigl(\frac{1}{\sqrt{n_r}}
    + \frac{1}{\sqrt{n_s}}\bigr)
    \text{polylog}\left(\frac{d}{w_{\min}}\right) \, .
  \] Further, it can be shown that $\ops{A-C}$ is
  $O\left(\sigma_{\max}\sqrt{n}\cdot
  \text{polylog}\Big(\frac{d}{w_{\min}}\Big)\right)$ with high probability
  (see~\cite{dasgupta2007spectral}).  Thus we conclude that, with high
  probability
  \[
    \ltwos{\mu_r - \mu_s} 
    \ge c\sqrt{km_0}\left(\frac{\ops{A-C}}{\sqrt{n_r}}
    + \frac{\ops{A-C}}{\sqrt{n_s}}\right).
  \] Thus the active separation requirement is satisfied. The proof for the
  inactive separation condition is similar. Finally, the proximity condition
  follows from the concentration properties of Gaussians.  \tododon{For the
  camera ready show this thoroughly; skipping for now.}
\end{proof}

\newpage
\section{Experimental Details}
\label{app:datasets}

\subsection{Datasets}

For all experiments involving real data, we use the EMNIST, FEMNIST, and
Shakespeare datasets. These datasets and their corresponding models are
available at the LEAF benchmark: \url{https://leaf.cmu.edu/}.  For client
selection experiments, we manually partition a subset of FEMNIST (first 10
classes) by assigning 2 classes to each device. There are 500 devices in total.
Both the number of training samples across all devices and the number of
training samples per class within each device follow a power law. We use the
natural partition of Shakespeare where each device corresponds to a speaking
role in the plays of William Shakespeare. We randomly sample 109 users from the
entire dataset.  For personalization experiments, following~\citet{ghosh2020},
we use a CNN-based model with one hidden layer and 200 hidden units trained
with a learning rate of $0.01$ and $10$ local updates on each device. 

\subsection{Choosing $k$ Based on Separation}

\begin{figure}[h!]
    \centering
    \includegraphics[width=0.65\textwidth]{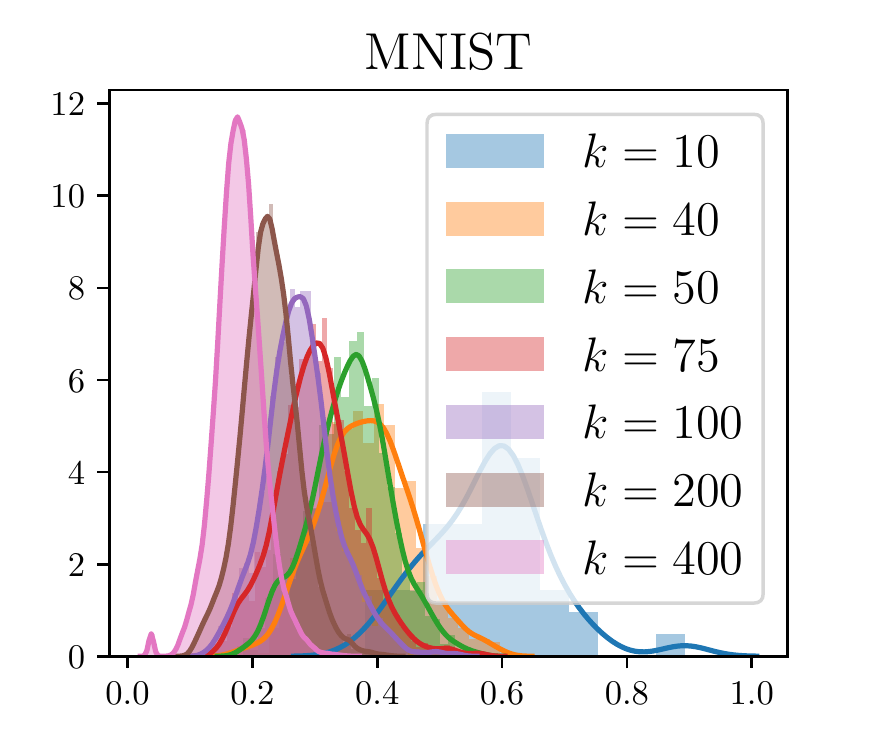}
    \caption{Distribution plot of $c_{rs}$, for various values of $k$ on the
    MNIST dataset. As can be seen, $c_{rs} < 1$ for most cluster pairs,
    indicating that the separation between them is relatively small.}
    \label{fig:mnist-sep}
\end{figure}

As mentioned in Section~4.2, to create our \textit{oracle clustering}, we
compute the quantity $c_{rs} = \frac{\norm{\mu_r -
\mu_s}}{2\sqrt{m_0}(\Delta_r + \Delta_s)}$ for each cluster pairs $(r,s)$,
for every candidate value of $k$ we are considering. We construct a
distribution plot of these $c_{rs}$. An example of such a plot
for the MNIST dataset is provided in Figure~\ref{fig:mnist-sep}. As can be seen
here, for all values of $k$, the relative separation is quite small. Thus even
for this oracle clustering, the actual separation between cluster means is
small.  To pick a $k$ for our oracle clustering, we pick a fixed value $c_0$
(say $0.5$) and then pick the value of $k$ which leads to maximum fraction of
cluster pairs $(r, s)$ to have $c_{rs} > c_0$.

\end{document}